\documentclass{article}

\usepackage{PRIMEarxiv}

\usepackage[utf8]{inputenc} 
\usepackage[T1]{fontenc}    
\usepackage{hyperref}       
\usepackage{url}            
\usepackage{booktabs}       
\usepackage{amsfonts}       
\usepackage{nicefrac}       
\usepackage{microtype}      
\usepackage{lipsum}
\usepackage{fancyhdr}       
\usepackage{graphicx}       

\usepackage{amsmath}
\usepackage{graphicx}
\PassOptionsToPackage{hyphens}{url}
\usepackage{hyperref}
\usepackage[algo2e,ruled,noline,noend]{algorithm2e}
\usepackage{subcaption}
\usepackage{multirow}
\usepackage{rotating}
\usepackage{xcolor}

\title{MemeTector: Enforcing deep focus for meme detection
}

\author{
  Christos Koutlis,
  Manos Schinas,
  Symeon Papadopoulos, 
  \\
  \texttt{ckoutlis@iti.gr, manosetro@iti.gr, papadop@iti.gr} \\
  CERTH-ITI, Greece\\
}

\begin{document}
\maketitle

\begin{abstract}
\looseness=-1
Image memes and specifically their widely-known variation \textit{image macros}, is a special new media type that combines text with images and is used in social media to playfully or subtly express humour, irony, sarcasm and even hate. It is important to accurately retrieve image memes from social media to better capture the cultural and social aspects of online phenomena and detect potential issues (hate-speech, disinformation). Essentially, the background image of an image macro is a regular image easily recognized as such by humans but cumbersome for the machine to do so due to feature map similarity with the complete image macro. Hence, accumulating suitable feature maps in such cases can lead to deep understanding of the notion of image memes. To this end, we propose a methodology, called \textit{Visual Part Utilization}, that utilizes the visual part of image memes as instances of the \textit{regular image class} and the initial image memes as instances of the \textit{image meme class} to force the model to concentrate on the critical parts that characterize an image meme. Additionally, we employ a trainable attention mechanism on top of a standard ViT architecture to enhance the model's ability to focus on these critical parts and make the predictions interpretable. Several training and test scenarios involving web-scraped regular images of controlled text presence are considered for evaluating the model in terms of  robustness and accuracy. The findings indicate that light visual part utilization combined with sufficient text presence during training provides the best and most robust model, surpassing state of the art. Source code and dataset are available at \url{https://github.com/mever-team/memetector}.
\end{abstract}


\section{Introduction}
\begin{figure}
     \centering
     \begin{subfigure}[b]{0.21\textwidth}
         \includegraphics[height=0.8\textwidth]{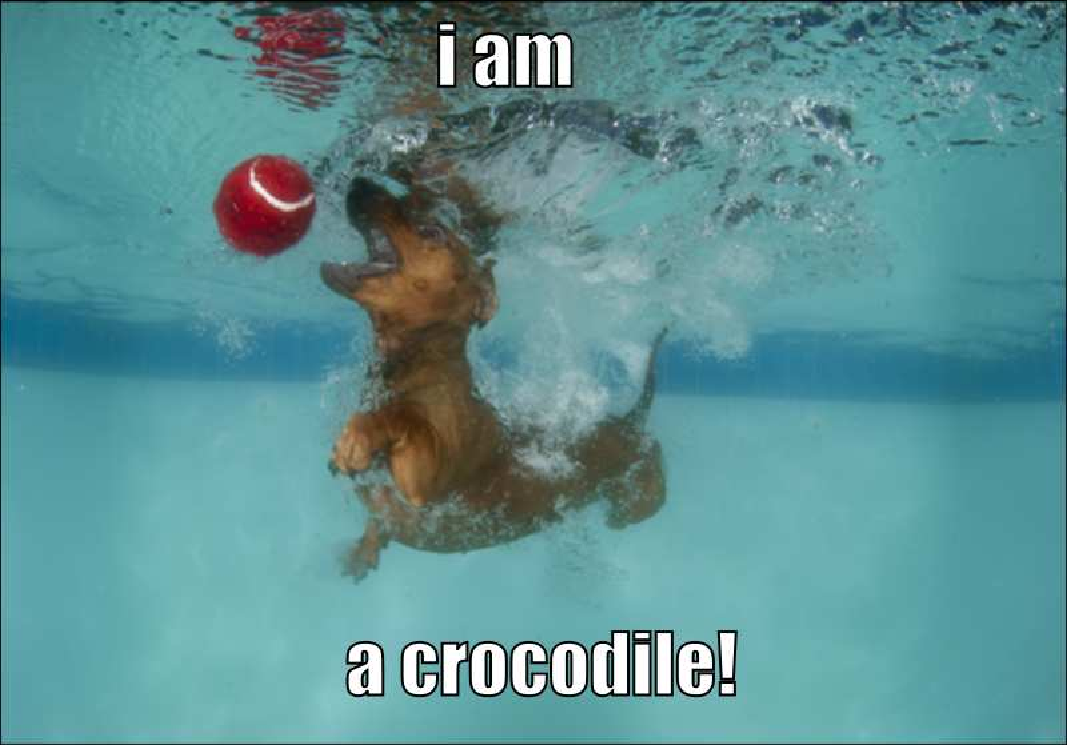}
         \caption{Image meme}
     \end{subfigure}
     \quad\quad
     \begin{subfigure}[b]{0.21\textwidth}
         \includegraphics[height=0.8\textwidth]{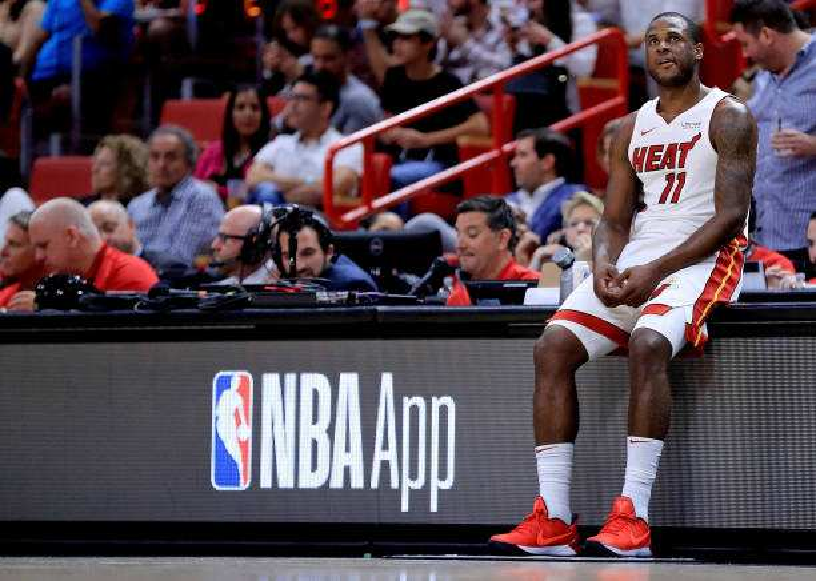}
         \caption{Regular image}
     \end{subfigure}
    
    \caption{Example image meme vs. a regular image.}
    \label{fig:meme_vs_regular}
\end{figure}

\textit{Image memes} have been established during the last years as a popular means of communication in social media. Their typical form, known as \textit{image macros}\footnote{\url{https://en.wikipedia.org/wiki/Image_macro}}, comprise images with overlay text at the top and/or bottom and are principally used to express a spectrum of concepts and emotions such as humour, irony, sarcasm and even hate. Memes and regular images have critical visual differences that render their discrimination an easy task for a human, such as the overlay text with a specific type of font size, color, family and position as well as the background image usually having a cultural reference or being memorable. 
In contrast, regular images potentially depict anything without certain constraints. In Figure~\ref{fig:meme_vs_regular}, we exemplify one \textit{image meme}\footnote{Source: Facebook's HateFul Memes dataset \cite{kiela20}} and one \textit{regular image}\footnote{Source: Google's Conceptual Captions dataset \cite{sharma18}} to showcase the differences between the two types of digital media.

Other forms of Internet memes also exist, for instance they might be plain text \cite{he19}, tweet screenshots, social statement cards, logos \cite{whatisameme}  or images reusing memorable visual elements in different creative ways such as Bernie Sanders' mittens\footnote{\url{https://en.wikipedia.org/wiki/Bernie_Sanders_mittens_meme}}. In addition, the adoption of different meme forms seems to be highly platform-dependent as community-specific vernaculars determine different meme cultures \cite{whatisameme}. Here, we only address the detection of the typical Internet meme form, namely image macros being a combination of background images with superimposed text as Figure~\ref{fig:meme_vs_regular}a.

In the framework of analysing digital social behavior and trends, image memes have attracted research interest \cite{segev15,zannettou18,xie11,dancygier17}, mostly with a focus on deep learning models for image meme classification \cite{afridi21,amalia18,smitha18,gaurav20} and more frequently for the detection of \textit{hateful} image memes \cite{aggarwal21,kiela21,zhou21,khedkar21}. The latter works utilize datasets with image memes and appropriate labeling, thus they do not put effort on detecting whether an input image is a meme or not. 
The detection of image memes and their discrimination from regular images is currently still a relatively understudied topic, and there are only few attempts in this direction \cite{miliani20,sinha19}.

\begin{figure*}[t]
    \centering
    \includegraphics[trim={0 10cm 0 0}, clip, width=\textwidth]{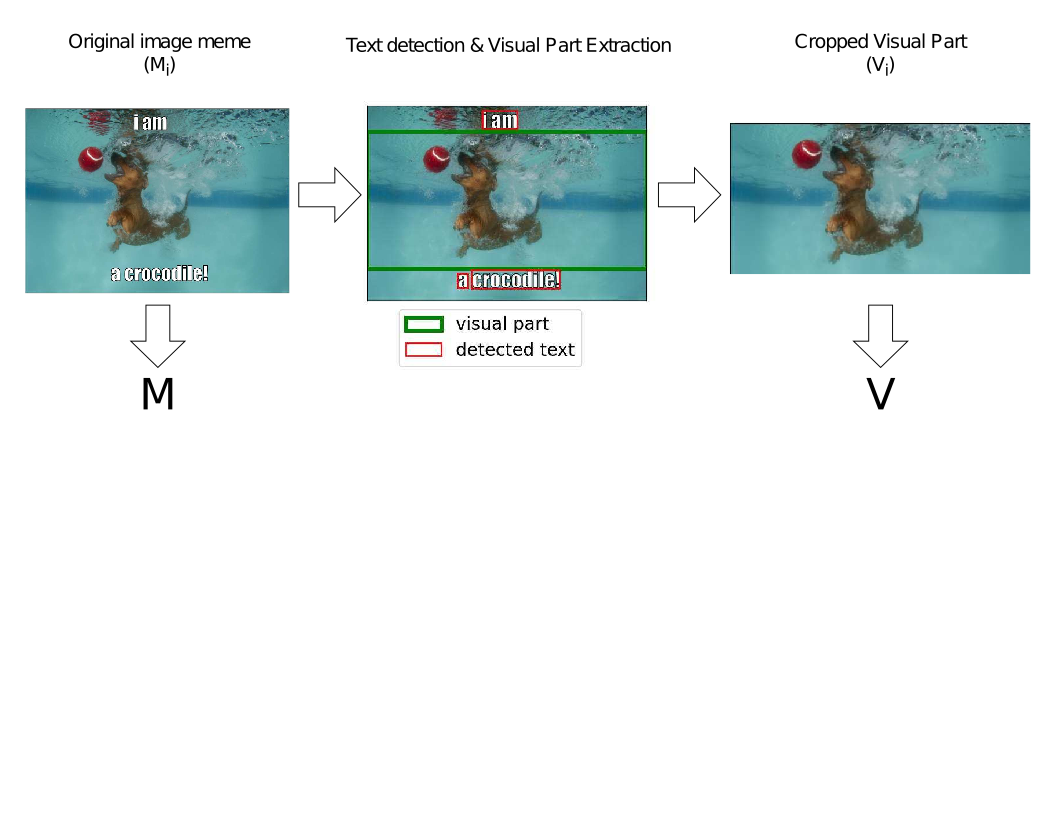}
    \caption{The proposed Visual Part Utilization process. The original image $M_i$, which belongs to the set of image memes $\textbf{M}$, is passed through the visual part extraction algorithm that identifies the corresponding visual part $V_i$, crops it and adds it to the set $\textbf{V}$. Best viewed in color.}
    \label{fig:VPU}
\end{figure*}

In this work, we present \textit{MemeTector}, a model for efficiently classifying images as memes or regular ones. Utilizing this model in online social environments to retrieve memes can facilitate the monitoring and analysis of web trends and behaviors as well as the detection of harmful practices that are sometimes carried out with the use of memes, such as hate-speech and disinformation. To enforce the model's focus on critical visual cues that characterize both classes, we propose Visual Part Utilization (VPU), a methodology for artificial dataset creation from an existing image meme dataset \cite{kiela20} and a deep learning architecture called ViTa, employing a trainable attention mechanism on top of a Vision Transformer (ViT) \cite{dosovitskiy20}. 
Although we propose ViTa for meme detection it is  general and can potentially be utilized to other tasks. Regarding VPU, from each image meme instance $M_i$ of the initial dataset we extract the largest part that contains no text and utilize it as a regular image instance. We denote the latter with $V_i$ and call it visual part of image meme $M_i$. In Figure~\ref{fig:VPU}, we present the construction process for set $\textbf{M}$ of image memes and $\textbf{V}$ of visual parts. In essence, a meme's background image is a regular image which is easily discriminated from the meme by humans while neural networks initially produce almost identical feature maps for both. VPU is thus considered here for effectively enhancing the learning of class distribution subtle peculiarities.

The paper is structured as follows.  Section~\ref{sec:relatedwork} reviews the related literature. Section~\ref{sec:method} elaborates the proposed methodology. Section~\ref{sec:setup} describes the experimental setup.  Section~\ref{sec:results} presents the results. Section~\ref{sec:conclusions} concludes the paper.

\section{Related work}\label{sec:relatedwork}
Previous related studies focus mainly on classification of memes into categories such as hateful or offensive \cite{suryawanshi20}. Due to their multi-modal nature the classification of image memes is most frequently treated as a multi-modal analysis problem by processing both visual and text content  \cite{afridi21}. Other studies attempt to classify image memes in terms of their sentiment (positive, negative, neutral) and type of humour (e.g. sarcastic or motivational) \cite{shrestha20}. However, the topic of this study precedes the work of classifying image memes into certain categories. First, one needs to know if an image is a meme or just a regular image before further analysing it. 

The topic of image meme detection, i.e. automatically discriminating image memes from regular images, has not yet received considerable attention from the research community. To our knowledge, there is only one dataset for meme detection, namely the DankMemes dataset \cite{miliani20}, which was released in 2020 but is publicly unavailable at the timing of writing the paper. DankMemes contains 2000 images related to the 2019 Italian government crisis, half of which are memes and the rest are regular images. In terms of competing approaches, only few meme detection methods have been recently presented  \cite{fiorucci20,setpal20,vlad20}. Also, the similar problem of identifying satire images on social media is addressed in \cite{sinha19}. Finally, a few other attempts exist on the Internet, for instance in blog posts or GitHub repositories, but are not peer reviewed.\footnote{\url{https://github.com/matyasbohacek/meme-detection}}\textsuperscript{,}\footnote{\url{https://medium.datadriveninvestor.com/memes-detection-android-app-using-deep-learning-d2c65347e6f3}}

This is the first paper to utilize the visual part of image memes as instances of the regular images class, thus enabling: (i) a 100\% accurate albeit automatic annotation process through the usage of already existing meme classification datasets, as well as (ii) the creation of a dataset of 40,000 images (x20 larger than the DankMemes dataset). Additionally, we are the first to employ a supplemental attention mechanism on top of a ViT architecture that empowers deep focus through the combination of different levels of information granularity and interpretability of the results.

\section{Methodology}\label{sec:method}
Here we present MemeTector's building blocks, namely VPU and ViTa.

\subsection{Visual Part Utilization}\label{subsec:vpu}

\subsubsection{Extraction}
To extract the visual part $V_i$ of a given image meme $M_i$ ($i=1,\dots,k$) one first needs to locate the text elements in it. To this end, we consider a state-of-the-art deep learning-based text detection model, called TextFuseNet \cite{Ye20}. This  processes $M_i$ and produces the set $\textbf{B}_i$ of detected text bounding boxes. We only keep boxes corresponding to whole words since the bounding boxes of letters are not useful for our task. Then, we apply Algorithm \ref{alg:VPU} to find the largest rectangle that contains no text and consider that part of $M_i$  as its visual part $V_i$.

A rectangle $R$ covers a fraction $p$ of the initial image area, namely $A_{R}=p\cdot W\cdot H$, where $W$ and $H$ are width and height of the initial image. Thus, $R$ can have width $\frac{\sqrt{p}\cdot W}{r}$ and height $\sqrt{p}\cdot H \cdot r$ to consider different aspect ratios $r$ and preserve the rectangle's area $A_R$. Given $p$, we can determine the upper and lower bounds of $r$ based on the image size i.e. $\frac{\sqrt{p}\cdot W}{r}\le W$ and $\sqrt{p}\cdot H\cdot r\le H$ that entail $\sqrt{p}\le r\le \frac{1}{\sqrt{p}}$. Similarly, given $p$ and $r$ we can determine the upper and lower bounds of the rectangle's center $c_R=(f_W\cdot W, f_H\cdot H)$:
\begin{equation}
    f_W\cdot W - \frac{W\sqrt{p}}{2r}\ge 0
\end{equation}
\begin{equation}
    f_W\cdot W + \frac{W\sqrt{p}}{2r}\le W
\end{equation}
\begin{equation}
    f_H\cdot H - \frac{\sqrt{p}\cdot H\cdot r}{2}\ge 0
\end{equation}
\begin{equation}
    f_H\cdot H + \frac{\sqrt{p}\cdot H\cdot r}{2}\le H
\end{equation}
that entail $\frac{\sqrt{p}}{2r}\le f_W\le 1-\frac{\sqrt{p}}{2r}$ and $\frac{r\sqrt{p}}{2}\le f_H\le 1-\frac{r\sqrt{p}}{2}$.

Consequently, we consider 17 equidistant values for $p$ and 10 equidistant values for $r$ , $f_W$ and $f_H$ covering the corresponding ranges. So, $17,000$ rectangles $R$ are created per $M_i$ from which we initially select the non-overlapping with any $B\in \textbf{B}_i$, and out of those select the ones with the maximum area. Finally, we randomly chose one rectangle $R_{V_i}$ and crop the corresponding part from $M_i$ in order to get its visual part $V_i$.

\begin{algorithm2e*}[t]
\caption{Visual part extraction}\label{alg:VPU}
\SetKwInOut{Input}{input}
\SetKwInOut{Output}{output}
\Input{$M_i$}
\Output{$V_i$}
$W,H \gets size(M_i)$\;
$\textbf{B}_i \gets \text{TextFuseNet}(M_i)$\;
$\mathcal R \gets \emptyset$\;
\For{$p\in [0.1,0.9]$, $r\in \big[\sqrt{p},\frac{1}{\sqrt{p}}\big]$, $f_W\in \big[\frac{\sqrt{p}}{2r}, 1-\frac{\sqrt{p}}{2r}\big]$, $f_H\in \big[\frac{r\sqrt{p}}{2}, 1-\frac{r\sqrt{p}}{2}\big]$}{
    $R \gets \big(f_W\cdot W - \frac{W\sqrt{p}}{2r}, f_H\cdot H - \frac{\sqrt{p}\cdot H\cdot r}{2}, f_W\cdot W + \frac{W\sqrt{p}}{2r}, f_H\cdot H + \frac{\sqrt{p}\cdot H\cdot r}{2}\big)$\;
                \If{$R\cap B=\emptyset$, $\forall B\in \textbf{B}_i$}{
                    $\mathcal R \gets \mathcal R \cup \{(R,p)\}$\;
                }
    }

$p_{max} \gets \max\Big(\big\{p\mid (R,p)\in\mathcal R\big\}\Big)$\;
$R_{V_i} \gets RandomSample\Big(\big\{R\mid \big((R,p)\in\mathcal R\big) \land \big(p=p_{max}\big)\big\}\Big)$\;
$V_i \gets crop(M_i,R_{V_i})$\;
\end{algorithm2e*}

\subsubsection{Utilization}\label{subsubsec:utilization}

We utilize the extracted visual parts $V_i$ of image memes $M_i$ as regular image instances in order to force the model's focus on the critical parts that discriminate them. More precisely, we consider the set $\textbf{M}=\{M_i\}_{i=1}^k$ that contains image memes and the set $\textbf{V}=\{V_i\}_{i=1}^k$ that contains the corresponding visual parts. In Figure~\ref{fig:VPU}, we illustrate the construction process of $\textbf{M}$ and $\textbf{V}$ through an example. To assess the extent to which VPU is useful we also conduct experiments mixing instances of $\textbf{V}$ and web-scraped regular images for the construction of regular images class $\textbf{R}$. Additionally, given the inherent text presence in image memes another crucial aspect to consider is the extent to which text presence in regular images affects the model's robustness. Hence, we consider two more sets as pools for $\textbf{R}$ construction, namely $\textbf{R}_p=\{R_i^p\}_{i=1}^k$ for web-scraped regular images with text presence and $\textbf{R}_a=\{R_i^a\}_{i=1}^k$ for web-scraped regular images with text absence. The model's objective is to correctly classify the instances of the two sets, $\textbf{M}$ and $\textbf{R}$, with:
\begin{equation}\label{eq:regular}
    \textbf{R}=\{V_i\}_{i=1}^{k\cdot (1-P_{W})}\cup \{R_i^p\}_{i=1}^{k\cdot P_{W}\cdot P_{T}}\cup \{R_i^a\}_{i=1}^{k\cdot P_{W}\cdot (1-P_{T})}
\end{equation}
where $P_W$ and $P_T$ denote the fraction of web-scraped regular images out of the total number of regular images and the fraction of web-scraped regular images with text presence out of the total number of web-scraped regular images, respectively. In that way, $\textbf{M}$ and $\textbf{R}$ preserve the same cardinality $k$.

\subsection{Model architecture}
We propose Vision Transformer with Trainable Attention (ViTa), that augments ViT \cite{dosovitskiy20} by an attention module leveraging information from multiple processing stages. This approach was first successfully tested on CNNs \cite{jetley18}.
\subsubsection{ViT}
The input image $\textbf{x}\in\mathbb{R}^{H\times W\times C}$ is reshaped into a sequence of flattened 2D patches $\textbf{x}_p\in\mathbb{R}^{N \times (P^2\cdot C)}$, where $H$ is height, $W$ is width, $C$ is number of channels, $P$ is patches' side length and $N=HW/P^2$ is number of patches. Then, $\textbf{x}_p$ is linearly projected to $D$ dimensions through a dense layer, a learnable \textit{class} token is added to the sequence and learnable 1D position embeddings are added to the $N+1$ tokens, resulting in the Transformer encoder inputs $\textbf{z}_0\in\mathbb{R}^{(N+1)\times D}$. 
Consequently, $L$ Transformer encoder layers process the inputs to produce the final vector representations:
\begin{equation}
    \textbf{z}_l^{'}=\text{MSA}(\text{LN}(\textbf{z}_{l-1}))+\textbf{z}_{l-1}
\end{equation}
\begin{equation}
    \textbf{z}_l=\text{MLP}(\text{LN}(\textbf{z}_l^{'}))+\textbf{z}_l^{'}
\end{equation}
where MSA is Multiheaded Self-Attention with $h$ heads \cite{vaswani17}, LN is Layer Normalization \cite{ba16}, MLP is a Multilayer Perceptron with two GELU \cite{hendrycks16} activated layers of $2\cdot D$ and $D$ number of units respectively and $l=1,\dots,L$. Finally, a general representation $\textbf{y}$, describing the whole image is extracted by passing $\textbf{z}_L^0$, namely the \textit{class} token's embedding after $L$ Transformer encoder layers, through Layer Normalization:
\begin{equation}
    \textbf{y}=\text{LN}(\textbf{z}_L^0)
\end{equation}

\subsubsection{Attention module}
ViT contains multiple self-attention layers in which the \textit{class} token's embedding receives information from the patch embeddings of the same layer. However, it lacks an attention module that combines information from past layers that capture semantics of different granularity levels. To this end, we first compute a compatibility score between $\textbf{y}$ and the patch embeddings of odd layers $\{\textbf{z}_1^{1:N}\}$, $\{\textbf{z}_3^{1:N}\}$, $\dots$, $\{\textbf{z}_n^{1:N}\}$ ($n=L-1$ if $L$ is even and $n=L$ if $L$ is odd), by:
\begin{equation}
    s_l^i = <\textbf{v},[\textbf{y};\textbf{z}_l^i]>
\end{equation}
where $i\in\{1,\dots,N\}$, $l\in\{1, 3, \dots, n\}$, $[\cdot;\cdot]$ denotes concatenation and $\textbf{v}\in\mathbb{R}^{2D}$ is a learnable vector. The attention weights are calculated through softmax as:
\begin{equation}
    a_l^i=\frac{\text{exp}(s_l^i)}{\sum_{i=1}^N \text{exp}(s_l^i)}
\end{equation}
and the context vectors are simply the weighted average of the corresponding layer's patch embeddings $\textbf{c}_l=\sum_{i=1}^N a_l^i\cdot\textbf{z}_l^i$.

\begin{figure*}
     \centering
     \begin{subfigure}[b]{0.25\textwidth}
         \centering
         \includegraphics[height=0.8\textwidth]{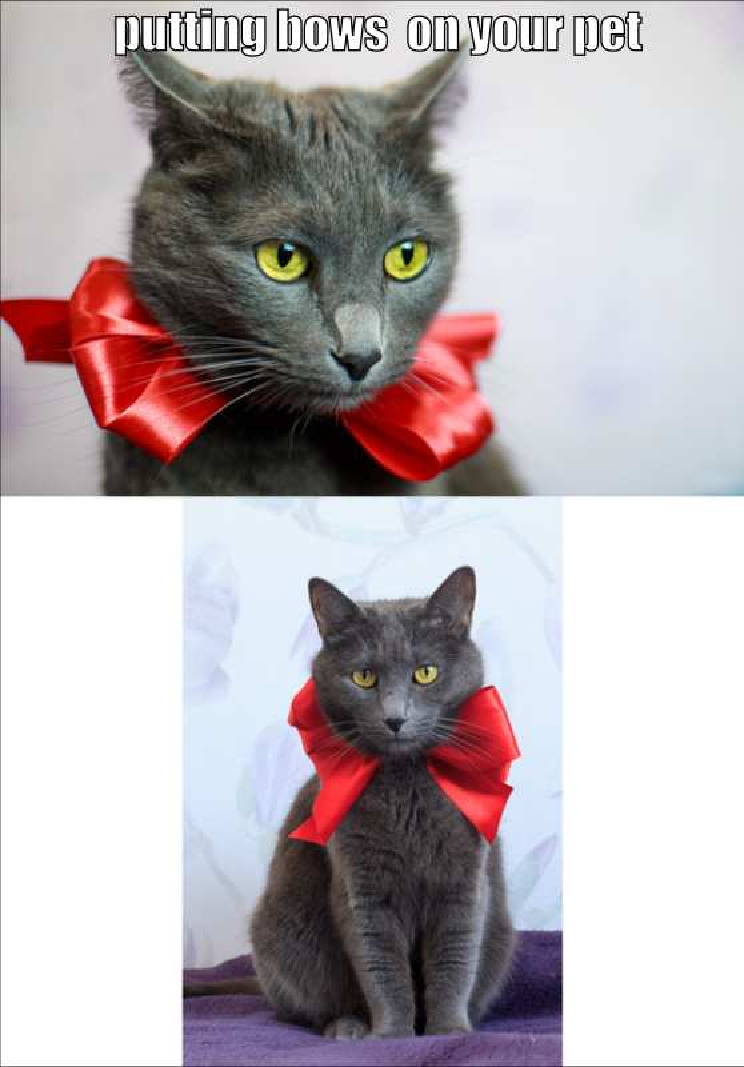}
         \caption{}
     \end{subfigure}
     \quad
     \begin{subfigure}[b]{0.25\textwidth}
         \centering
         \includegraphics[height=0.8\textwidth]{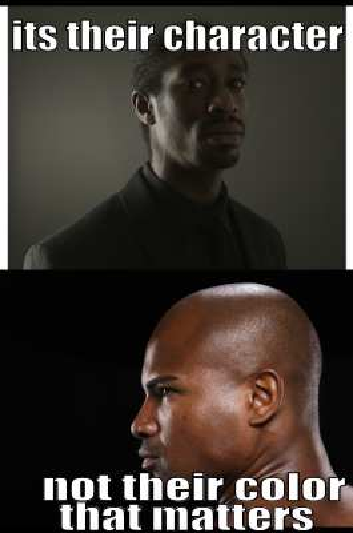}
         \caption{}
     \end{subfigure}
     \quad
     \begin{subfigure}[b]{0.2\textwidth}
         \centering
         \includegraphics[height=0.8\textwidth]{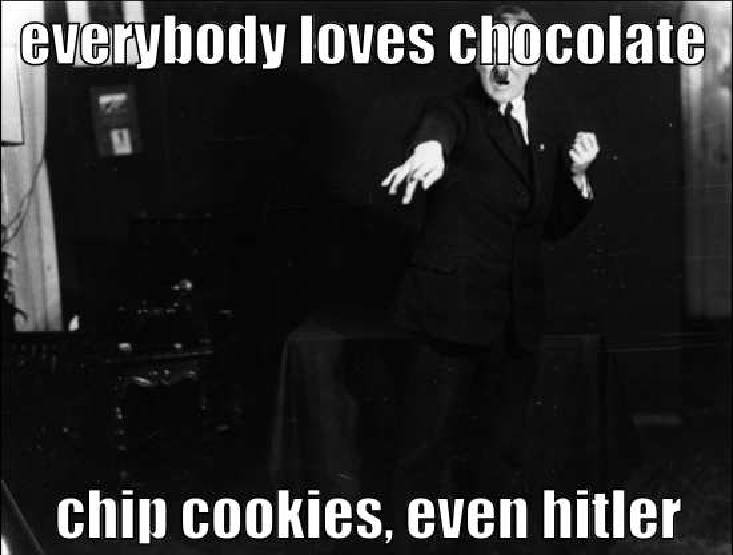}
         \caption{}
     \end{subfigure}
    
    \begin{subfigure}[b]{0.25\textwidth}
         \centering
         \includegraphics[width=\textwidth]{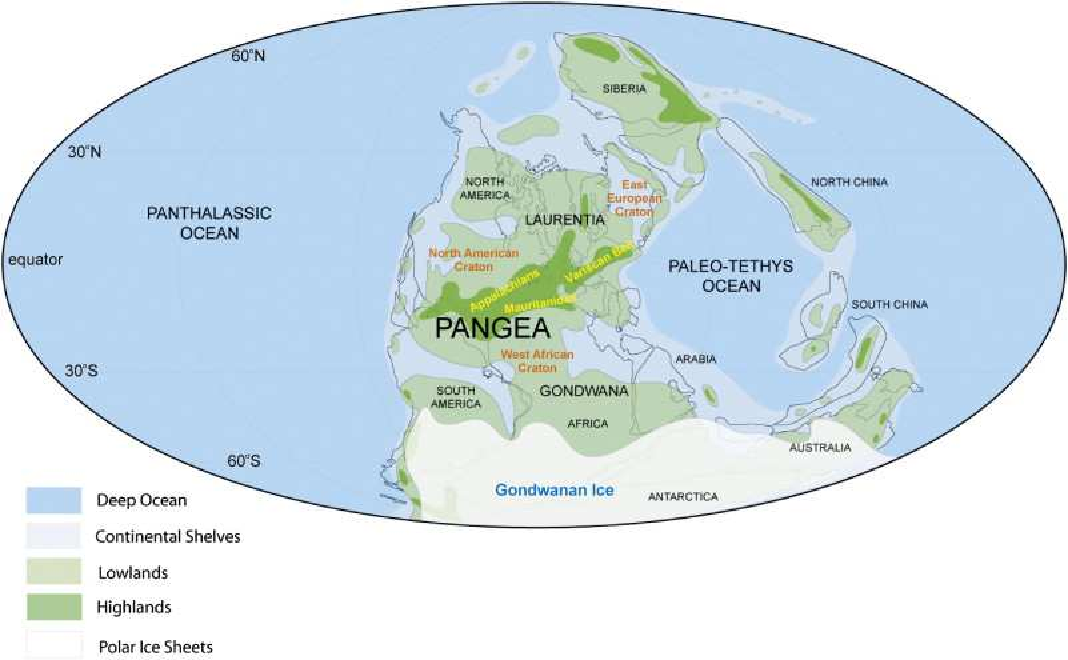}
         \caption{}
     \end{subfigure}
     \quad
     \begin{subfigure}[b]{0.25\textwidth}
         \centering
         \includegraphics[height=0.8\textwidth]{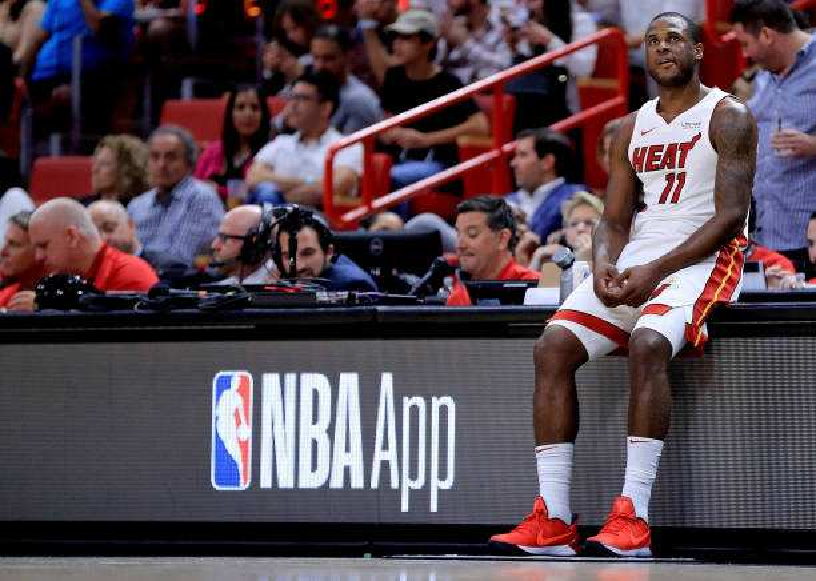}
         \caption{}
     \end{subfigure}
     \quad
     \begin{subfigure}[b]{0.25\textwidth}
         \centering
         \includegraphics[height=0.8\textwidth]{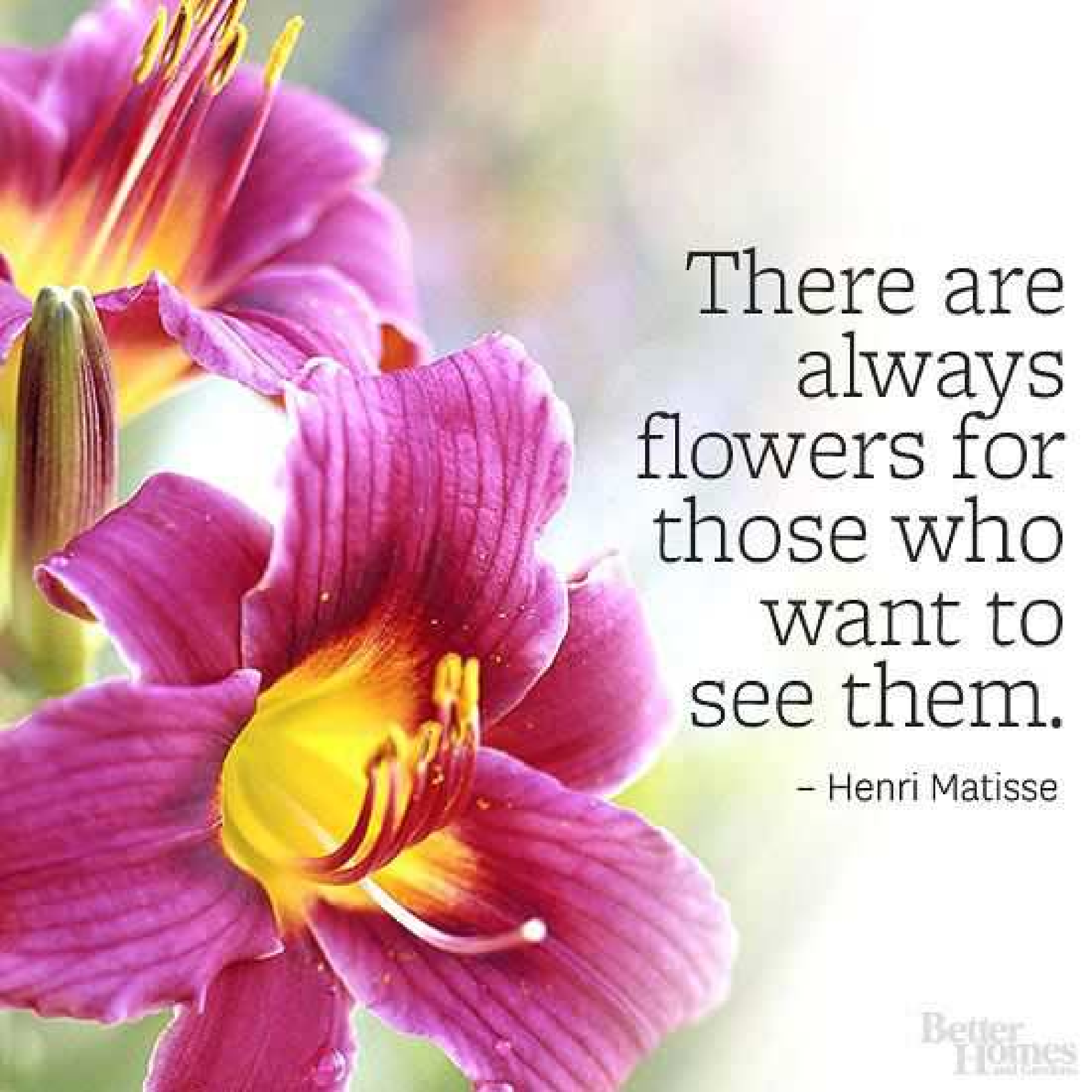}
         \caption{}
     \end{subfigure}
     
     \begin{subfigure}[b]{0.25\textwidth}
         \centering
         \includegraphics[height=0.8\textwidth]{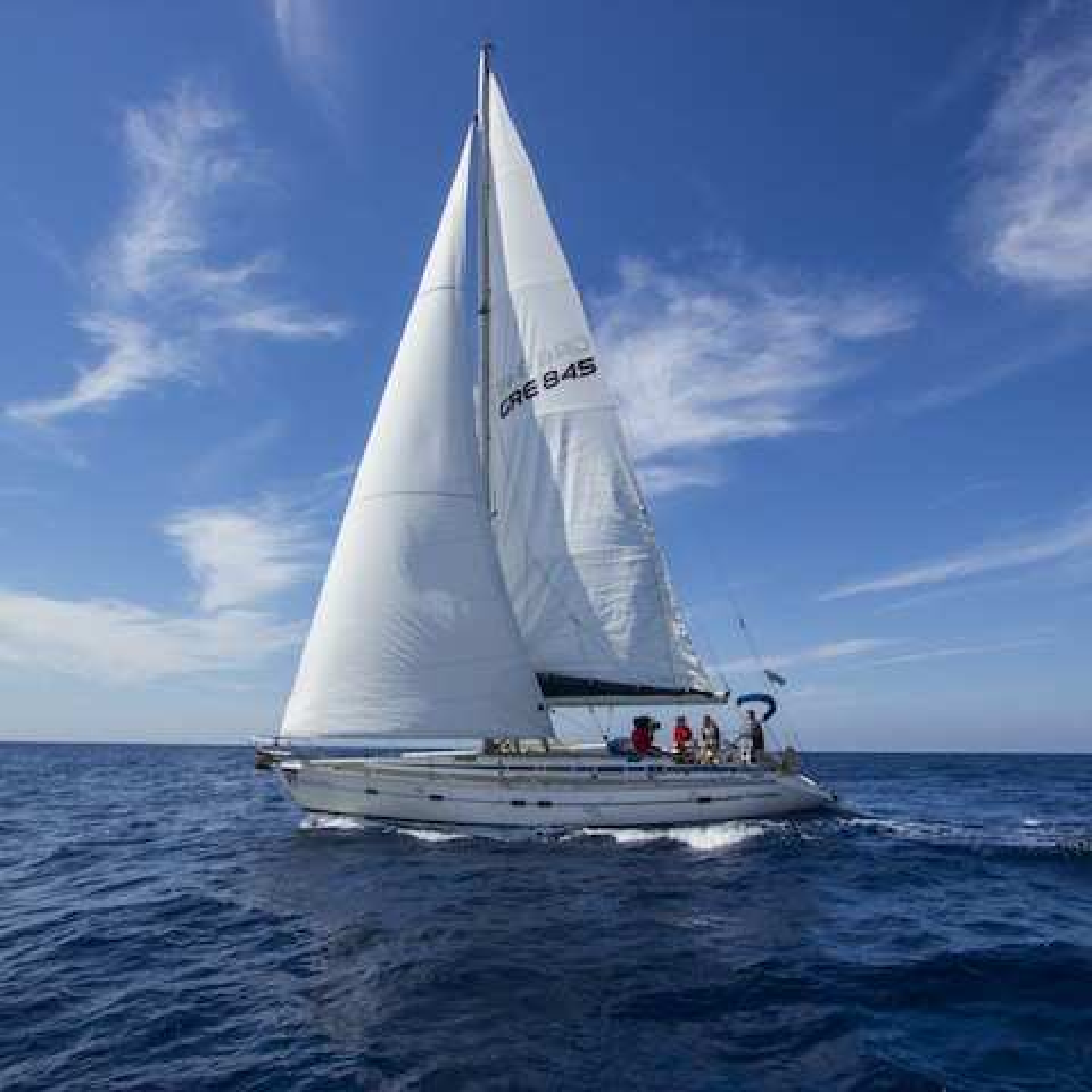}
         \caption{}
     \end{subfigure}
     \quad
     \begin{subfigure}[b]{0.25\textwidth}
         \centering
         \includegraphics[height=0.8\textwidth]{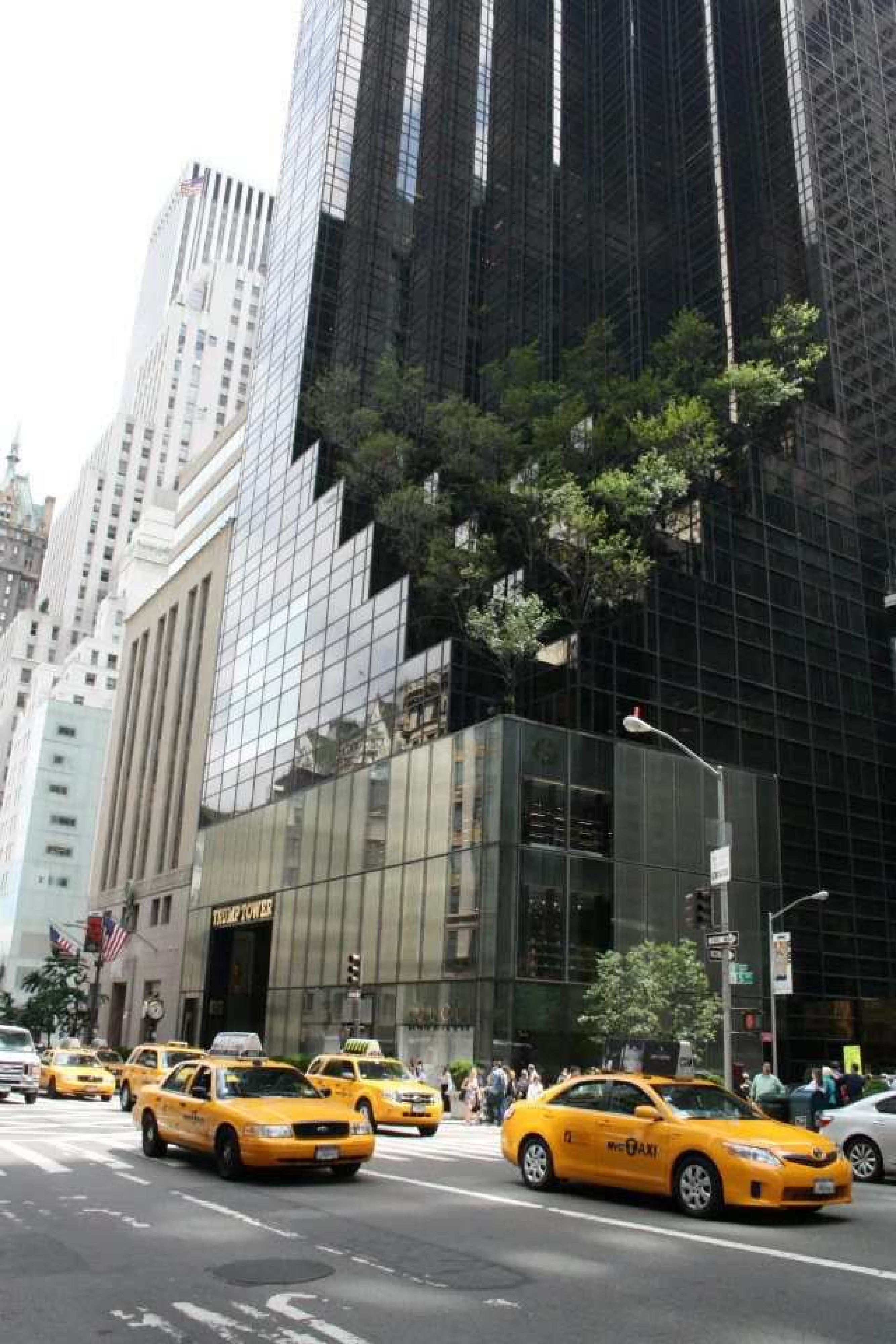}
         \caption{}
     \end{subfigure}
     \quad
     \begin{subfigure}[b]{0.25\textwidth}
         \centering
         \includegraphics[height=0.8\textwidth]{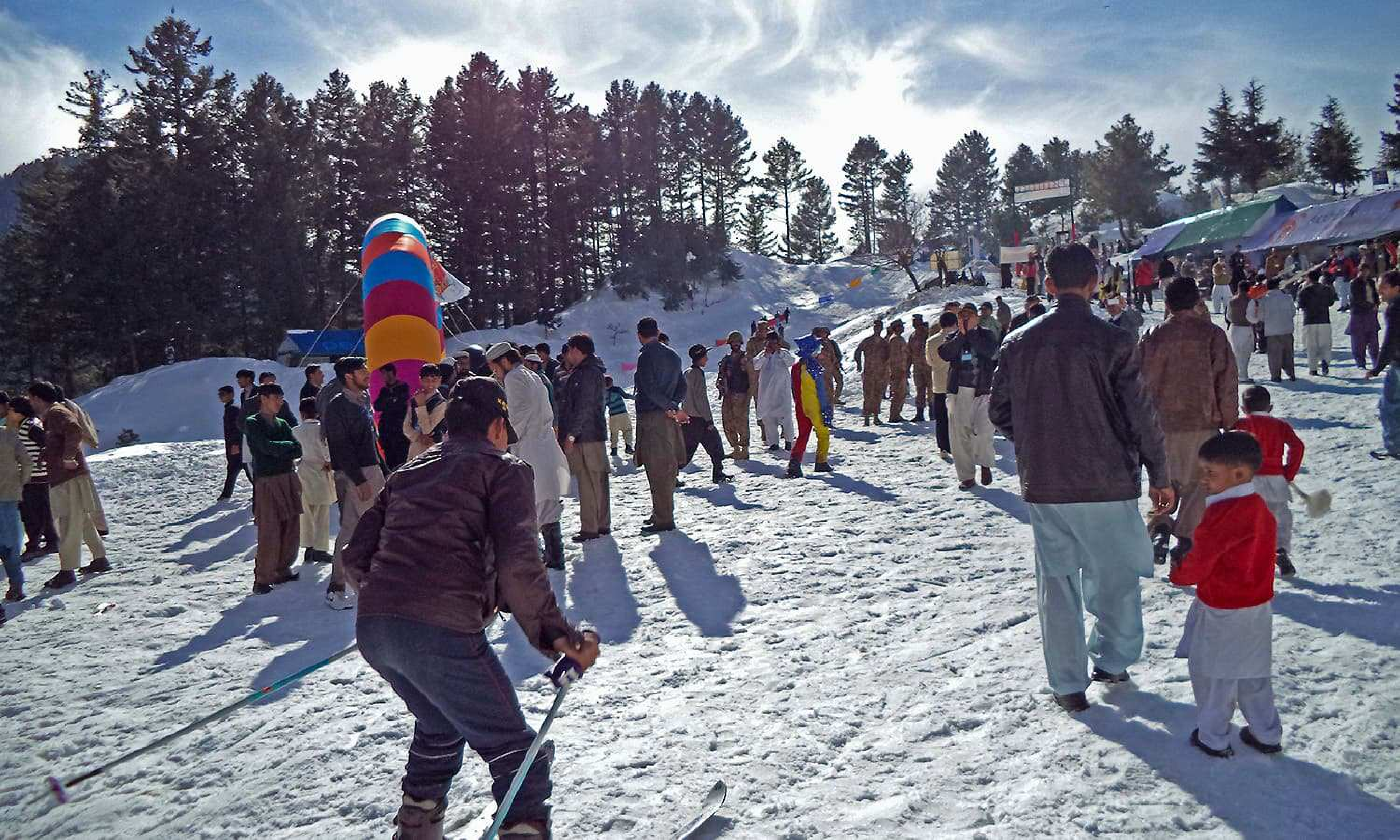}
         \caption{}
     \end{subfigure}
    \caption{Example images from the Hateful Memes (a, b, c) and the Google's Conceptual Captions with (d, e, f) and without (g, h, i) text presence.}
    \label{fig:datasets}
\end{figure*}

\subsubsection{Classification module}
The concatenation of all context vectors $\textbf{c}= [\textbf{c}_1;\textbf{c}_3;\dots;\textbf{c}_n]\in\mathbb{R}^{D^{'}}$ ($D^{'}=([n/2]+1)\cdot D$, where $[\cdot]$ denotes integer part) is processed by three dense layers for the final prediction, the first two are GELU activated and the last has one sigmoid unit:

\begin{equation}\label{eq:modeloutput}
   y=\text{sigmoid}(\textbf{w}_3\cdot\text{GELU}(\textbf{w}_2\cdot\text{GELU}(\textbf{w}_1\cdot \textbf{c})))
\end{equation}
where $\textbf{w}_1\in\mathbb{R}^{2048\times D^{'}}$, $\textbf{w}_2\in\mathbb{R}^{1024\times 2048}$, and $\textbf{w}_3\in\mathbb{R}^{1024}$.

\section{Experimental setup}\label{sec:setup}

\subsection{Datasets}

To form a suitable dataset that can be used for the task of meme detection, we merge instances from existing datasets that contain image memes and regular images respectively.

For the image meme class $\textbf{M}$, we consider the Hateful Memes Dataset \cite{kiela20}. This is a multimodal dataset for hateful meme detection containing 10,000 image memes. 
We do not take into account the nature of these memes, being hateful or not, in our analysis rather we use them all under the class of image memes.  Figure~\ref{fig:datasets} (a)-(c) illustrates three indicative examples.

For the regular images class $\textbf{R}$, apart from the VPU methodology, explained in Section~\ref{subsec:vpu}, that we apply on the Hateful Memes dataset to obtain $\textbf{V}$, we also consider part of the widely-used web-scraped Google's Conceptual Captions dataset \cite{sharma18}. Specifically, we randomly sample images in order to construct $\textbf{R}_p$ and $\textbf{R}_a$ (see Section~\ref{subsubsec:utilization}), with the text presence property automatically assessed through TextFuseNet \cite{Ye20} if it detects at least one text instance. Figure~\ref{fig:datasets} presents three indicative examples with ((d)-(f)) and without ((g)-(i)) text presence.

\subsection{Sample mixing and splitting}
As a starting point for the image meme class $\textbf{M}$, we consider the 10,000 instances of the Hateful Memes dataset. Then, we extract the visual parts of these image memes resulting in 9,984 images considered as regular to form the set $\textbf{V}$. The mean area fraction across all $V_i\in\textbf{V}$ is 64.3\%. For the remaining 16 images the VPU algorithm was unable to find a rectangle with no overlap with text. Thus, for all four sets $\textbf{M}$, $\textbf{V}$, $\textbf{R}_p$ and $\textbf{R}_a$ we  consider the same size of $k$=9,984 instances. To do so, we discard the same 16 image memes from $\textbf{M}$ and sample from Google's Conceptual Captions dataset $k$=9,984 instances to form $\textbf{R}_p$ and another $k$=9,984 instances to form $\textbf{R}_a$ respectively.

\begin{table*}[t]
\caption{Composition scenarios for regular images class ($\textbf{R}$) construction based on $P_W$ and $P_T$.}
\label{tab:scenarios}
\centering
\begin{tabular}{c|c|cccc|cccc|cccc}
\hline
$P_W$ &  0\% &\multicolumn{4}{c|}{33\%}&\multicolumn{4}{c|}{67\%}&\multicolumn{4}{c}{100\%}\\
\hline
$P_T$ & 0\% & 0\% & 33\% & 67\% & 100\% & 0\% & 33\% & 67\% & 100\%  & 0\% & 33\% & 67\% & 100\% \\
\hline
\end{tabular}
\end{table*}

For sample mixing, in order to form the class of regular images $\textbf{R}$ we only need to determine $P_W$ and $P_T$ (see Equation~\ref{eq:regular}). Also, to assess the impact of both VPU and text presence in the model's performance we consider several dataset composition scenarios $S_i$=$(P_W,P_T)$, with $i$=$1,\dots,13$, for $\textbf{R}$, presented in Table~\ref{tab:scenarios}.

Furthermore, we consider the same scenarios both on the training and test sets and experiment with \textit{crossed scenarios} $(S_i, S_j)$, e.g. the model is trained on $S_1$=($P_W$=0\%, $P_T$=0\%) but evaluated on $S_{13}$=($P_W$=100\%, $P_T$=100\%), resulting in 13$\cdot$13=169 crossed scenarios to analyse. For sample splitting, we initially select 85\% training, 5\% validation and 10\% test samples for each set $\textbf{M}$, $\textbf{V}$, $\textbf{R}_p$, $\textbf{R}_a$ and construct $\textbf{R}$ per split. For $\textbf{M}$ and $\textbf{V}$ we consider the same index split in order not to include the visual part $V_i$ in one split e.g. training, and the initial image meme $M_i$ in another, e.g. test. The training and validation sets always derive from the same scenario $S_i$, while the evaluation is performed on all test scenarios $S_j$.

\subsection{Competitive models}\label{subsec:competitive}
Meme detection is an understudied research topic. However, published studies exist such as leveraging ImageNet pre-trained ResNet features \cite{fiorucci20}, combining ResNet, AlexNet and DenseNet fine-tuned on the task \cite{setpal20}, and considering VGG16 and ResNet fine-tuned on the task \cite{vlad20}. These works consider text representations as additional inputs yet do not assess their impact. In this study, 
we only focus on the analysis of the visual signal for three reasons: 
\begin{enumerate}
    \item meme detection is a visually driven task as the critical information regarding existing text is the fonts' size, color, family and position rather than the actual text content
    \item reliably recognizing text in image memes (let alone regular images) is a challenging task on its own, so an error-prone text recognition component would add another layer of unnecessary complexity
    \item manual text recognition by humans, such as in the dataset used by the previously mentioned studies, is unrealistic for automatic meme detection.
\end{enumerate}
Hence, we consider state of the art and baseline models from the image classification domain as competitive methods. More precisely, we consider VGG16 \cite{simonyan14}, ResNet50 \cite{he16}, EfficientNetB5 \cite{tan19}, and ViT\footnote{For ViT we consider the same architecture as ViTa discarding only the attention module, which means that $\textbf{y}$ is directly passed to Equation~\ref{eq:modeloutput} replacing $\textbf{c}$. This of course changes the size of $\textbf{w}_1$ which for ViT belongs to $\mathbb{R}^{2048\times D}$.} \cite{dosovitskiy20}.

\subsection{Training details}
For ViT and ViTa, we consider input size $(H,W,C)$=$(250, 250, 3)$ and patch size $P$=25 that entail $N$=100, projection dimension $D$=64, $L$=8 Transformer encoder layers and $h$=4 MSA heads. In total ViTa has 3.4M parameters trained from scratch as we also do for ViT. We consider the pre-trained on ImageNet weights for EfficientNetB5, ResNet50 and VGG16, discard the last layer, add a dense sigmoid activated layer to all and train only this very last layer.

All models are trained for 20 epochs with batch size 64, using the AdamW optimizer \cite{loshchilov18} with weight decay 1e-3 and the binary cross-entropy loss function. The first 10\% of the iterations are linear warm-up steps with learning rate $\lambda(t)$ from 0 to 1e-3 which then decays as below:
\begin{equation}
    \lambda(t) = \frac{\text{1e-3}}{(1+d\cdot t\cdot 1.001^t)}
\end{equation}
where $t$ is the iteration and $d$=1e-3/20. We checkpoint our models based on validation accuracy. For ViT and ViTa, we preprocess input images first through normalizing to [0,1] and then through standardizing the values by subtracting the mean and dividing by the standard deviation per channel (computed on the training set). For the other models, we use the standard preprocessing pipeline provided by keras.

All models are evaluated on various test settings using the binary accuracy metric.

\section{Results}\label{sec:results}

\begin{sidewaystable*}
\begin{center}
\begin{minipage}{\textheight}
\caption{MemeTector's performance in terms of binary accuracy over all crossed scenarios. Best performance per test scenario is denoted with bold letters.}
\label{tab:accuracyall}

\begin{tabular}{cc|c|c|cccc|cccc|cccc}
\multirow{16}{*}{\rotatebox[origin=c]{90}{\textbf{Training scenario}}}&\multicolumn{15}{c}{\textbf{Test scenario}}\\
&\textbf{$P_W$}&&\textbf{0\%}&\multicolumn{4}{c|}{\textbf{33\%}}&\multicolumn{4}{c|}{\textbf{67\%}}&\multicolumn{4}{c}{\textbf{100\%}}\\
\cline{2-16}
&&\textbf{$P_T$}&\textbf{0\%} & \textbf{0\%} & \textbf{33\%}& \textbf{67\%} & \textbf{100\%} & \textbf{0\%} & \textbf{33\%}& \textbf{67\%} & \textbf{100\%} & \textbf{0\%} & \textbf{33\%}& \textbf{67\%} & \textbf{100\%}\\
\cline{2-16}
&\rotatebox[origin=c]{90}{\textbf{0\%}}&\textbf{0\%}&94.1 &93.9 &93.4 &93.1 &92.7 &93.6 &92.9 &92.1 &90.7 &93.1 &92.2 &90.7 &89.0\\
\cline{2-16}
&\multirow{4}{*}{\rotatebox[origin=c]{90}{\textbf{33\%}}}&\textbf{0\%}&96.8 &96.5 &95.5 &94.5 &94.2 &95.6 &93.9 &93.1 &91.9 &95.2 &92.9 &91.4 &89.6\\
&&\textbf{33\%}&96.5 &96.7 &95.7 &95.1 &95.0  &96.4 &95.1 &94.4 &94.0  &96.2 &94.8 &93.9 &92.7\\
&&\textbf{67\%}&96.4 &96.2 &95.7 &95.4 &95.2 &96.0  &95.5 &94.7 &93.9 &95.3 &94.6 &93.7 &92.3\\
&&\textbf{100\%}&96.6 &96.6 &96.2 &95.6 &95.4 &96.5 &95.4 &94.9 &94.3 &95.9 &94.9 &93.8 &92.6\\
\cline{2-16}
&\multirow{4}{*}{\rotatebox[origin=c]{90}{\textbf{67\%}}}&\textbf{0\%}&96.4 &96.5 &95.9 &95.5 &95.2 &96.2 &95.2 &94.7 &94.2 &96.2 &94.8 &93.9 &92.7\\
&&\textbf{33\%}&\textbf{97.8} &\textbf{97.4} &\textbf{96.8} &96.3 &96.2 &\textbf{97.3} &96.0  &95.9 &95.3 &96.7 &95.5 &94.7 &93.8\\
&&\textbf{67\%}&96.4 &96.3 &95.9 &95.7 &95.4 &96.4 &95.7 &95.0  &94.9 &96.3 &95.4 &94.5 &94.0 \\
&&\textbf{100\%}&97.2 &97.1 &\textbf{96.8} &\textbf{96.6} &\textbf{96.7} &96.6 &\textbf{96.1} &\textbf{96.3} &\textbf{95.8} &\textbf{97.0}  &\textbf{96.3} &\textbf{96.0}  &\textbf{95.5}\\
\cline{2-16}
&\multirow{4}{*}{\rotatebox[origin=c]{90}{\textbf{100\%}}}&\textbf{0\%}&94.5 &95.2 &94.5 &93.9 &93.3 &95.5 &94.3 &93.7 &92.4 &95.5 &93.9 &92.8 &91.4\\
&&\textbf{33\%}&96.0  &96.0  &95.8 &95.5 &95.3 &96.2 &95.9 &95.4 &95.2 &96.7 &95.8 &95.4 &94.7\\
&&\textbf{67\%}&94.5 &95.2 &95.0  &94.8 &94.7 &95.7 &95.5 &95.1 &94.7 &95.0  &95.2 &94.9 &94.\\
&&\textbf{100\%}&94.6 &94.9 &94.6 &94.7 &94.6 &95.2 &95.0  &94.8 &94.9 &95.1 &95.1 &94.8 &94.5\\
\cline{2-16}
\end{tabular}

\end{minipage}
\end{center}
\end{sidewaystable*}

\subsection{Ablation study}\label{subsec:crossed}

\begin{table*}[!ht]
    \footnotesize
    \caption{Impact of VPU usage in terms of aggregated accuracy. A($S_i$,$S_j$) denotes accuracy of the model when trained on $S_i$ and evaluated on $S_j$. In all cases 10$\leq$i,j$\leq$13.}
    \label{tab:vpu_comparison}
    \centering
    
    \begin{tabular}{c|c|cccc|c}
        \hline
        &VPU&$\max\limits_{i}\big($A($S_{i}$,$S_{10}$)$\big)$&$\max\limits_{i}\big($A($S_{i}$,$S_{11}$)$\big)$&$\max\limits_{i}\big($A($S_{i}$,$S_{12}$)$\big)$&$\max\limits_{i}\big($A($S_{i}$,$S_{13}$)$\big)$&$\sum\limits_{i,j}$A($S_i$,$S_j$)/16\\
        \hline
         VGG16&no&		 97.35 &94.69  &93.29 &92.49& 92.53  \\
         ResNet50&no&		 \textbf{98.05} &\textbf{96.30}  &95.90 &95.25 &94.24 \\
         EfficientNetB5&no&	 96.55  &94.04 &93.89 &93.99 &92.41\\
         ViT&no&			 96.00 &95.75 &95.20 &94.44  &94.24\\
         ViTa (ours)&no&			 96.70 &95.85 &95.40 &94.74 &94.69\\
         \hline
         && A($S_9$,$S_{10}$)& A($S_9$,$S_{11}$)& A($S_9$,$S_{12}$)& A($S_9$,$S_{13}$)&$\sum\limits_{i}$A($S_9$,$S_i$)/4\\
         \hline
         ViTa (ours)&yes&		 97.00 &\textbf{96.30}  &\textbf{96.00} &\textbf{95.50} &\textbf{96.20}\\
         \hline
    \end{tabular}
    
\end{table*}

In Table~\ref{tab:accuracyall}, we illustrate the performance of the proposed MemeTector model on all crossed scenarios between training and test settings. For easier interpretation of the results, we denote best performance per test scenario with bold letters.
First, we observe that in almost all crossed scenarios MemeTector obtains high accuracy scores ranging between 89.0 and 97.8 (mean 94.98 and standard deviation 1.47). The $S_9$=($P_W$=67\%, $P_T$=100\%) training scenario not only provides better performance on average and the highest accuracy values in most test scenarios (10 out of 13), but it preserves model robustness against all test scenarios as well. On the contrary, the highest variability in performance is observed in the $S_2$=($P_W$=33\%, $P_T$=0\%) training scenario that provides high accuracy in test scenarios with $P_T$=0\% yet low accuracy in test scenarios with text presence. The best performance is achieved in the ($S_7$, $S_1$) scenario, where MemeTector is trained on ($P_W$=67\%, $P_T$=33\%) and evaluated on ($P_W$=0\%, $P_T$=0\%). The worst performance is obtained with ($S_1$, $S_{13}$), where MemeTector is trained on ($P_W$=0\%, $P_T$=0\%) and evaluated on ($P_W$=100\%, $P_T$=100\%). This is expected, as feeding the model with samples of different nature makes it generalize better to easy scenarios such as $S_1$, while evaluating it on a dissimilar and harder setting compared to what it was trained on leads to lower performance. It is also remarkable that when evaluating on $S_{10}$-$S_{13}$ scenarios, training on $S_9$ outperforms the models trained on $S_{10}$-$S_{13}$. This fact showcases the usefulness of the VPU methodology in training set construction.

In Table~\ref{tab:vpu_comparison}, we present comparative results with regards to VPU. The first four columns show the maximum model performance when trained without the use of VPU, and the last row shows the corresponding results of MemeTector when trained using the VPU. Additionally, the last column of Table~\ref{tab:vpu_comparison} provides the average model performance. It is observed that VPU improves our model performance and also when incorporated, MemeTector performs better than competition in 3 out of 4 test scenarios as well as on average.

\subsection{Comparative study}\label{subsec:compare}
Based on the analysis presented in Section~\ref{subsec:crossed}, we employ our best configuration to compare with the competitive models presented in Section~\ref{subsec:competitive}. The most robust training scenario that also provides best performance in most evaluation scenarios is $S_9$=($P_W$=67\%, $P_T$=100\%). Hence, we consider this configuration for comparison with state of the art, namely the proposed MemeTector model architecture trained using 33\% VPU created regular images and 67\% web-scraped images that all contain text. Additionally, we train and evaluate VGG16, ResNet50, EfficientNetB5 and ViT on all crossed scenarios. 

In Table~\ref{tab:avg_acc}, we present the average model performance on all crossed scenarios where the proposed methodology achieves the highest score. Moreover,  Table~\ref{tab:surpass} presents the fraction of crossed scenarios where MemeTector surpasses the performance of competitive models. Although the average difference between our model and the second to best, namely ViT is only 0.81\%, MemeTector actually outperforms the latter in 87.57\% of the crossed scenarios. Similarly, the proposed methodology outperforms the baselines in the majority of cases.

\begin{minipage}{.45\linewidth}
      \captionof{table}{Models' performance in terms of average binary accuracy across all crossed scenarios.}
    \label{tab:avg_acc}
    \centering
        \begin{tabular}{cc}
    \hline
        model & accuracy \\
        \hline
         VGG16 & 91.36\% \\
         ResNet50 & 92.31\%\\
         EfficientNetB5 & 90.05\%\\
         ViT & 94.17\%\\
         MemeTector (ours)& \textbf{94.98\%}\\
         \hline
    \end{tabular}
\end{minipage}%
\quad\quad\quad\quad
\begin{minipage}{.45\linewidth}
      \captionof{table}{Fraction of crossed scenarios that the MemeTector model surpasses competitive models.}
    \label{tab:surpass}
    \centering
        \begin{tabular}{cc}
    \hline
        model & fraction \\
        \hline
         VGG16 & 147/169=86.98\% \\
         ResNet50 & 109/169=64.50\%\\
         EfficientNetB5 & 162/169=95.86\%\\
         ViT & 148/169=87.57\%\\
         \hline
    \end{tabular}
\end{minipage}

\begin{figure*}[!ht]
    \centering
    \includegraphics[width=\textwidth]{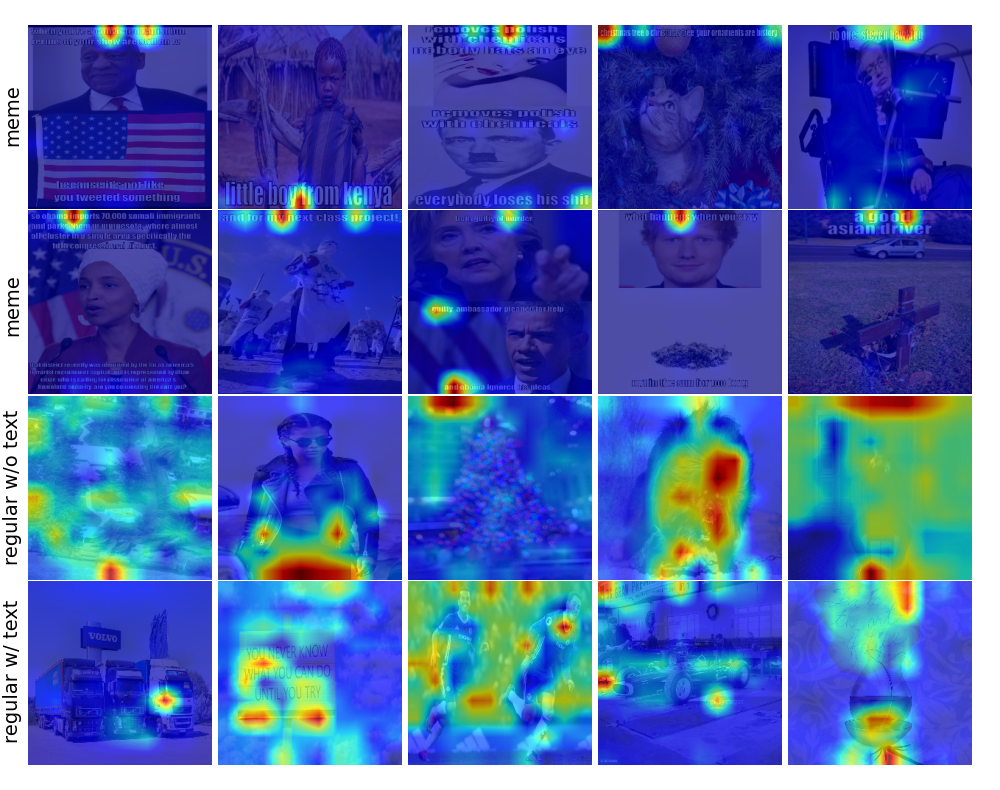}
    \caption{Attention plots from MemeTector's trainable attention mechanism. All predictions are correct. The upper 10 are image memes, while the lower 10 are regular images. From regular images the first 5 have text absence while the last 5 have text presence.}
    \label{fig:attention}
\end{figure*}

\subsection{Attention plots}\label{subsec:attentionplots}

\begin{figure*}[t]
     \centering
     \begin{subfigure}[b]{0.22\textwidth}
         \centering
         \includegraphics[width=\textwidth]{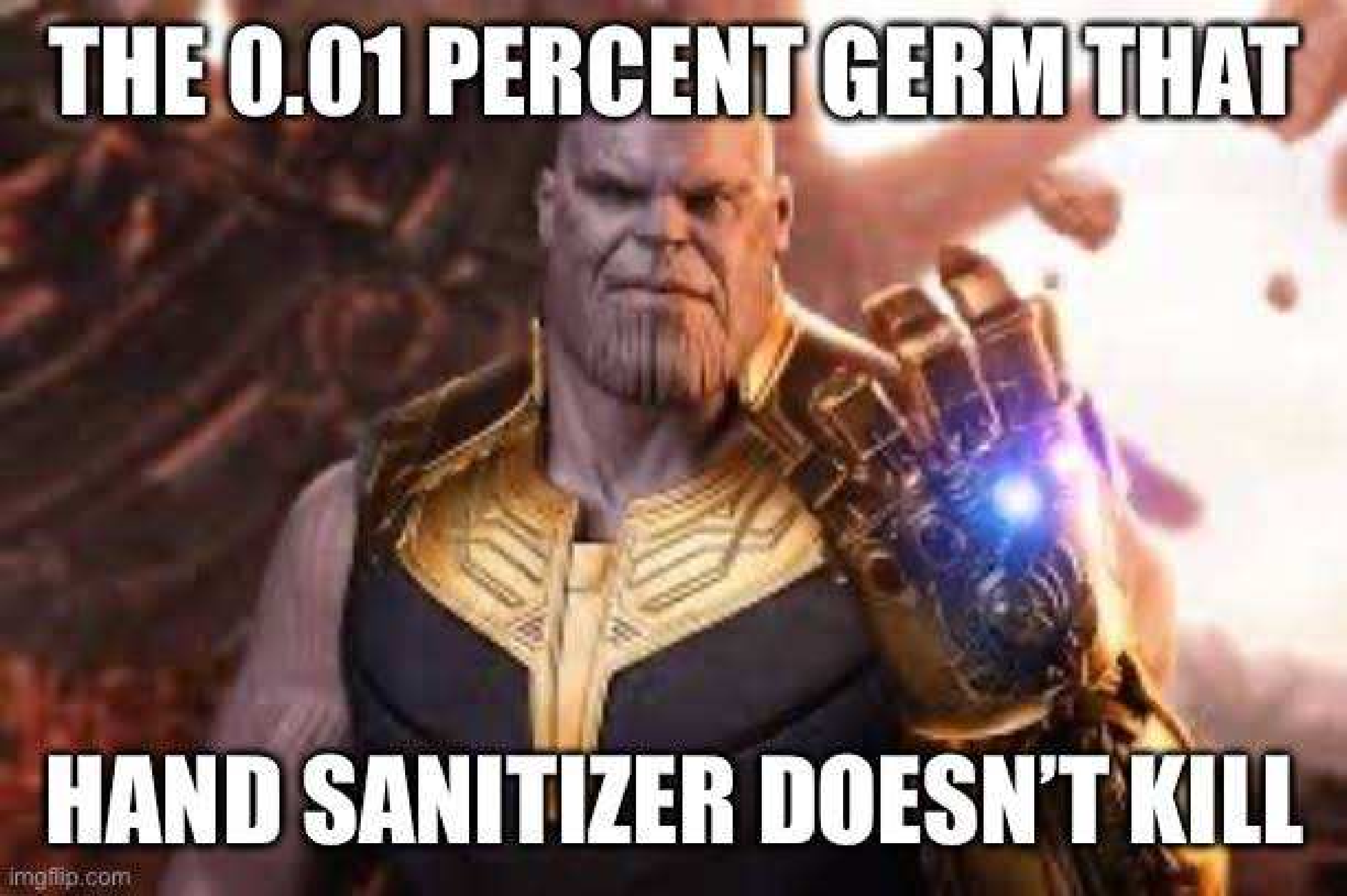}
         \caption{meme$\rightarrow$\textcolor{green!70!black}{meme}}
     \end{subfigure}
     \quad
     \begin{subfigure}[b]{0.22\textwidth}
         \centering
         \includegraphics[height=\textwidth]{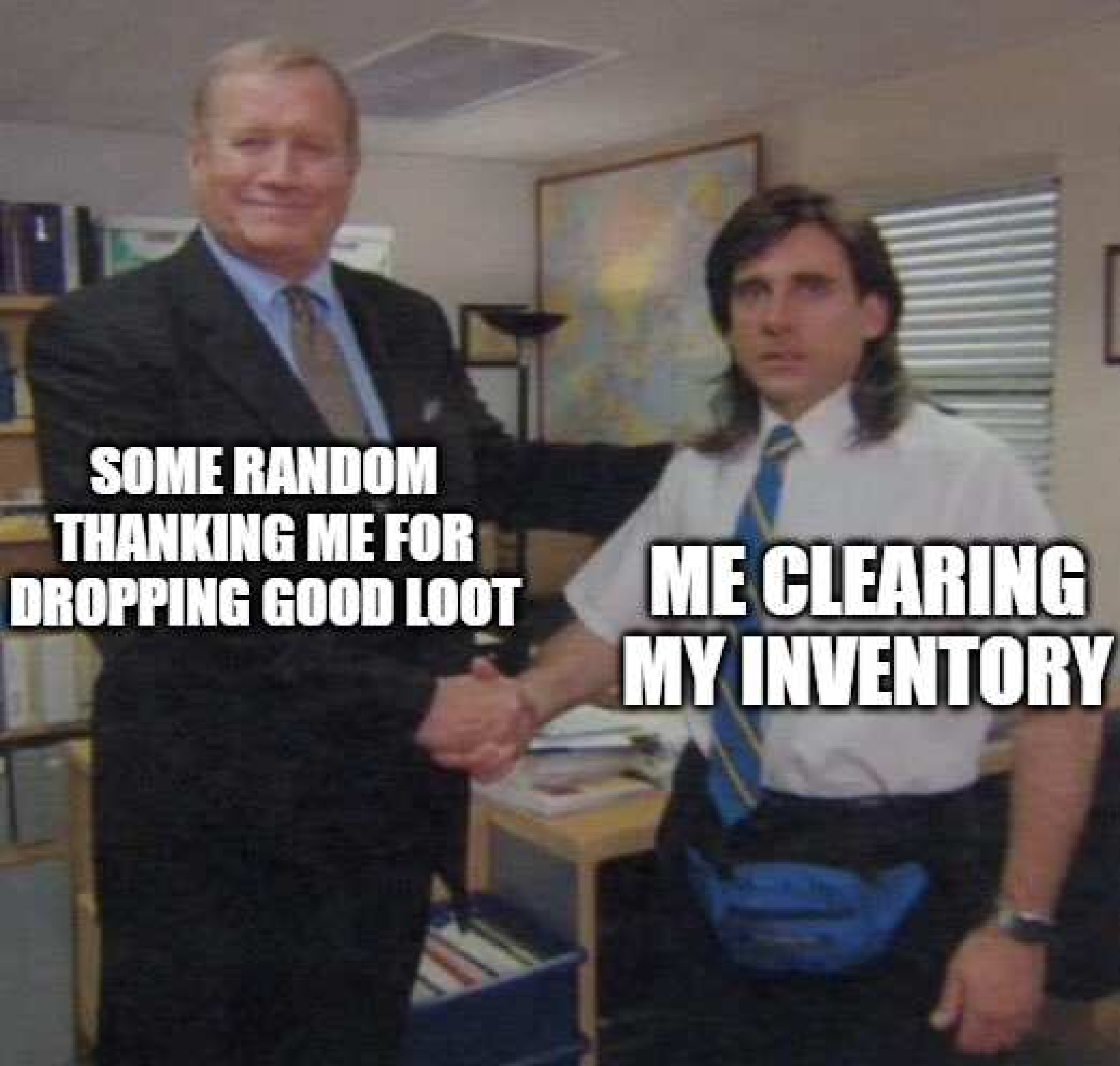}
         \caption{meme$\rightarrow$\textcolor{green!70!black}{meme}}
     \end{subfigure}
     \quad
     \begin{subfigure}[b]{0.22\textwidth}
         \centering
         \includegraphics[height=\textwidth]{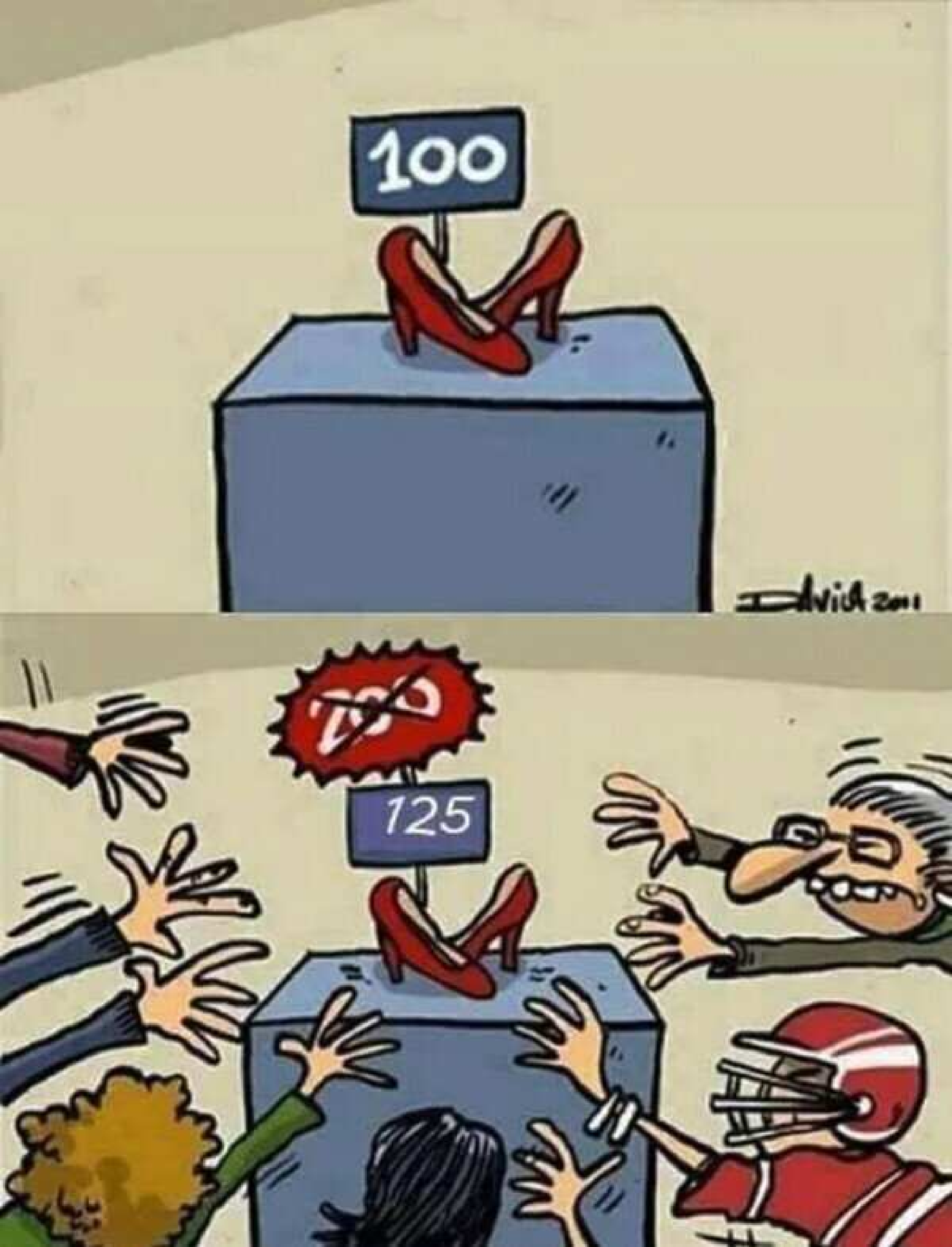}
         \caption{regular$\rightarrow$\textcolor{green!70!black}{regular}}
     \end{subfigure}
     \quad
     \begin{subfigure}[b]{0.22\textwidth}
         \centering
         \includegraphics[height=\textwidth]{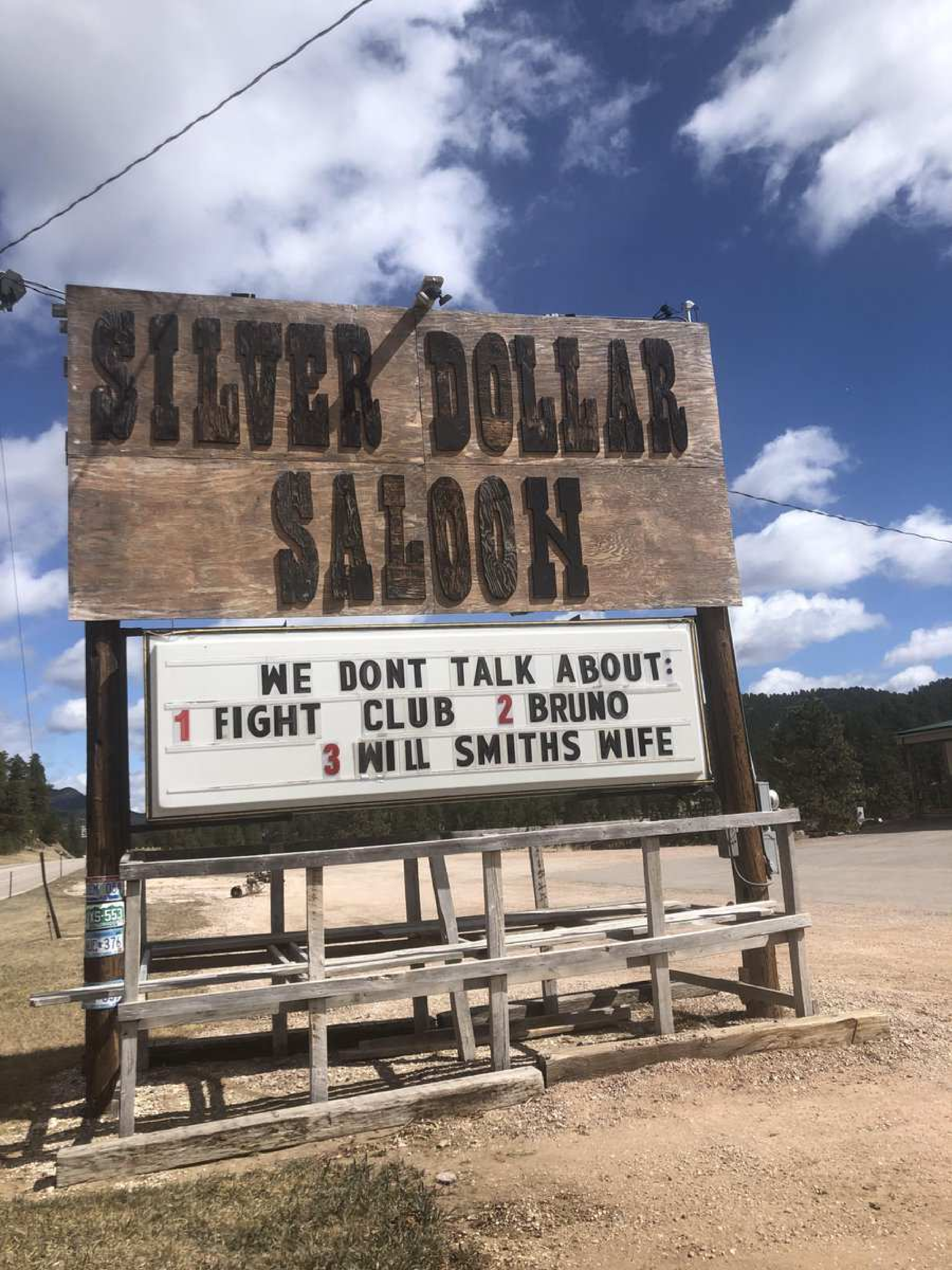}
         \caption{regular$\rightarrow$\textcolor{green!70!black}{regular}}
     \end{subfigure}
     
     \begin{subfigure}[b]{0.22\textwidth}
         \centering
         \includegraphics[height=\textwidth]{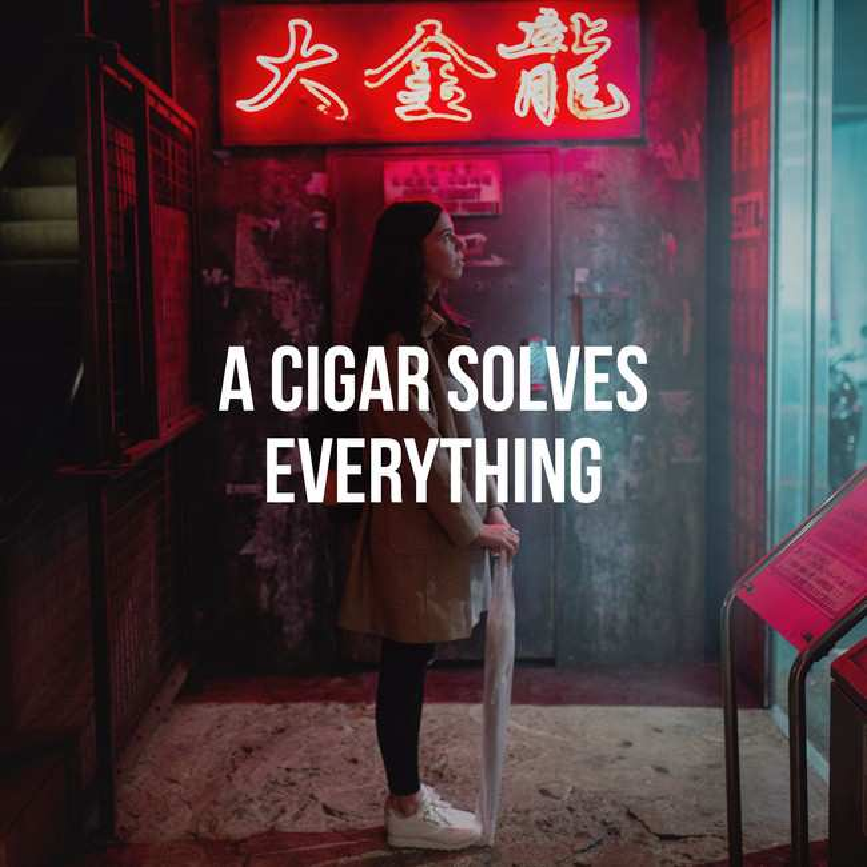}
         \caption{regular$\rightarrow$\textcolor{red}{meme}}
     \end{subfigure}
     \quad
     \begin{subfigure}[b]{0.22\textwidth}
         \centering
         \includegraphics[height=\textwidth]{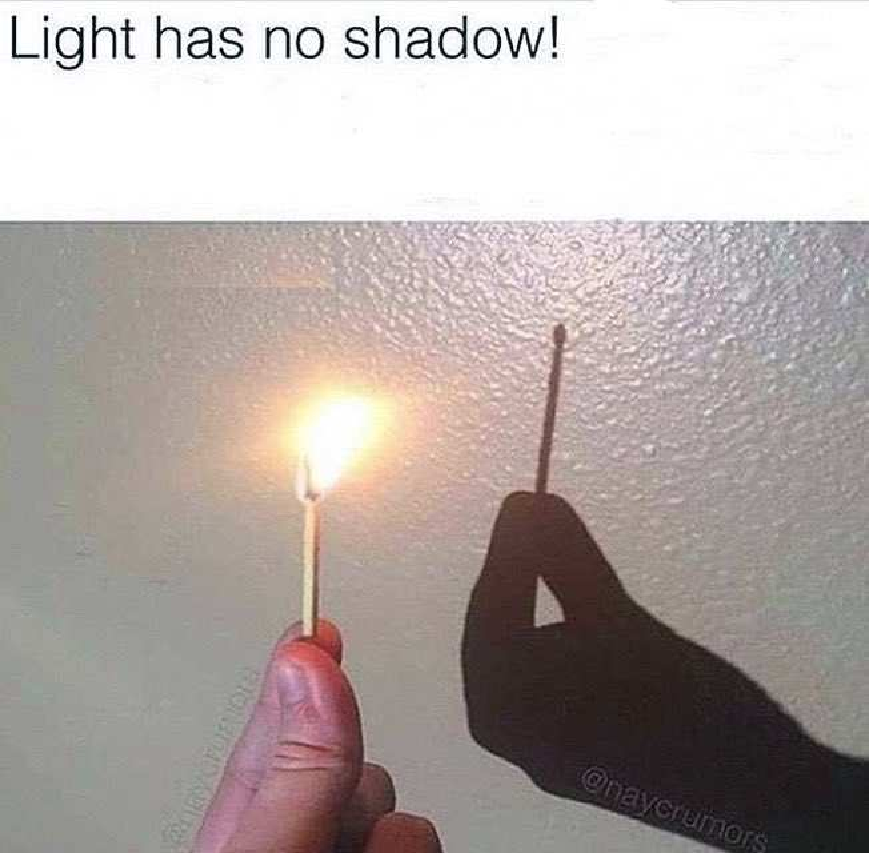}
         \caption{regular$\rightarrow$\textcolor{red}{meme}}
     \end{subfigure}
     \quad
     \begin{subfigure}[b]{0.22\textwidth}
         \centering
         \includegraphics[width=\textwidth]{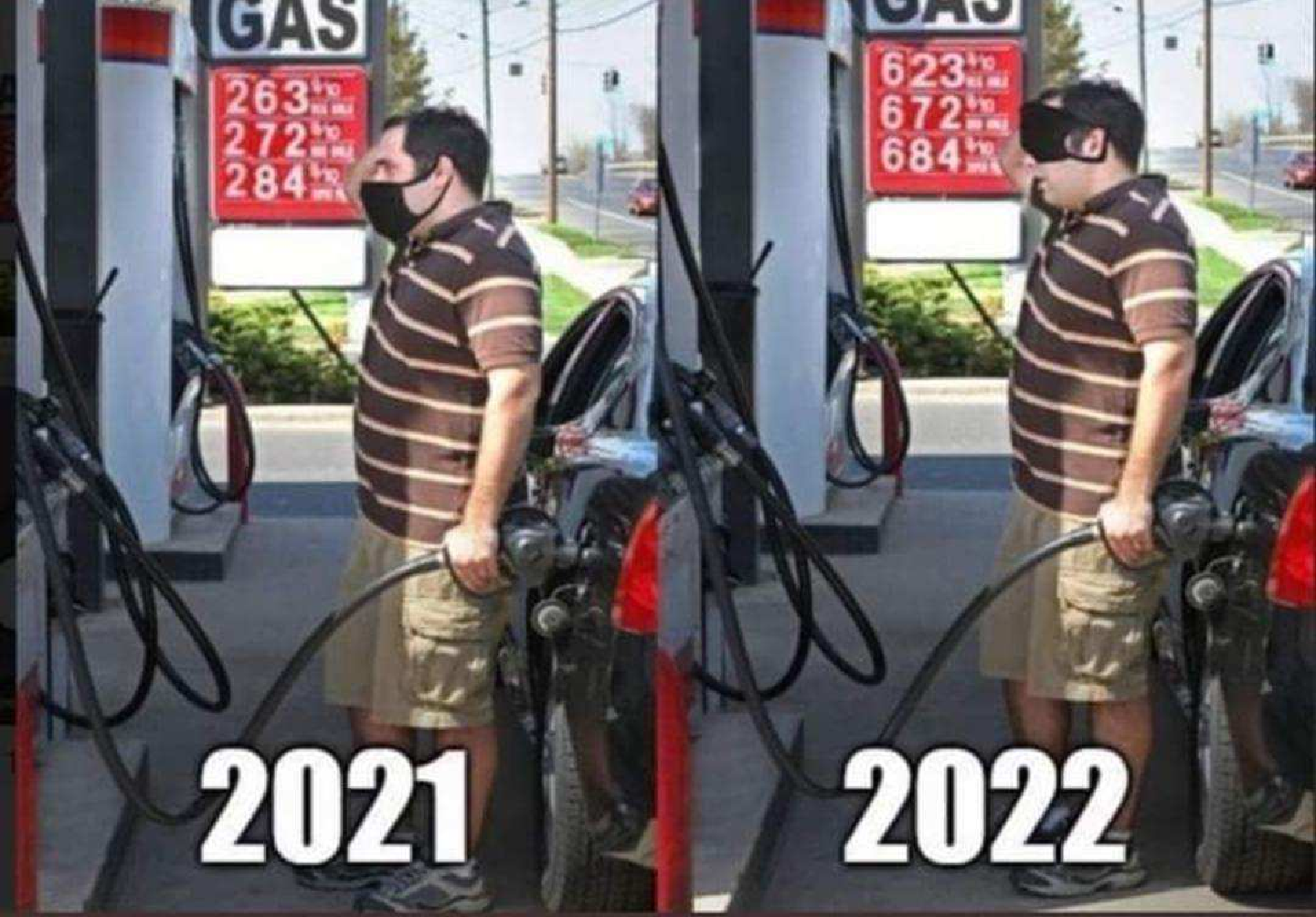}
         \caption{meme$\rightarrow$\textcolor{red}{regular}}
     \end{subfigure}
     \quad
     \begin{subfigure}[b]{0.22\textwidth}
         \centering
         \includegraphics[height=\textwidth]{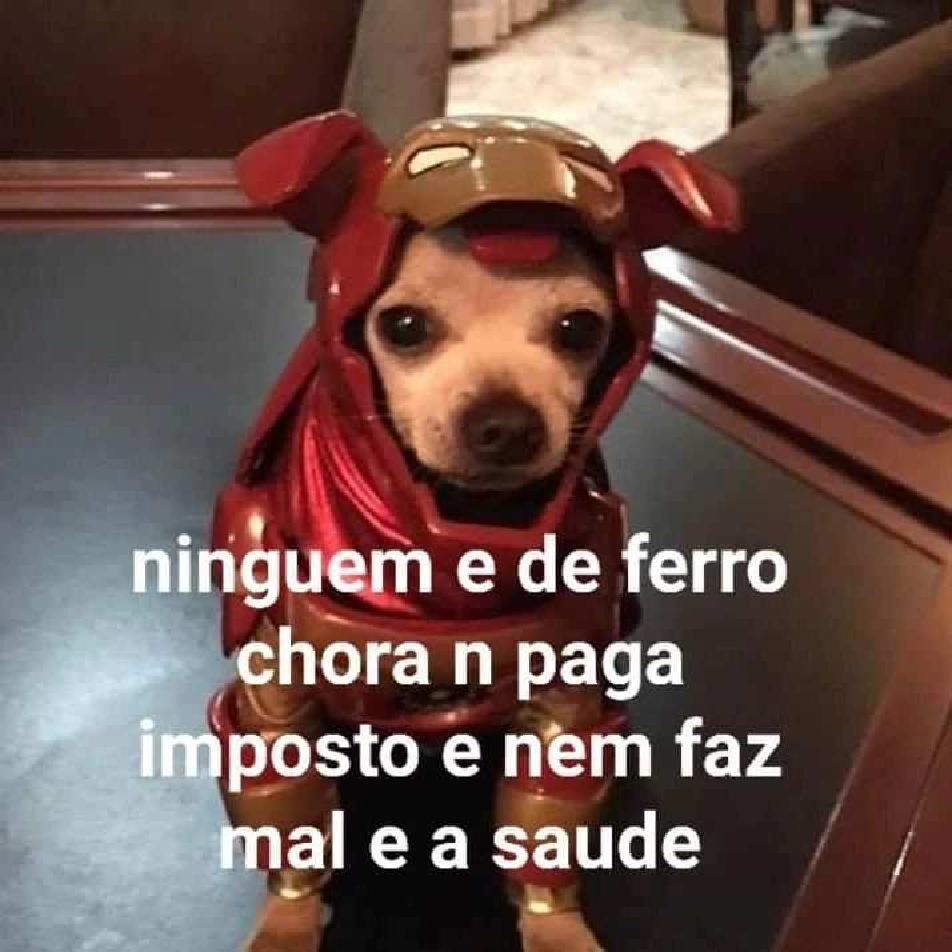}
         \caption{meme$\rightarrow$\textcolor{red}{regular}}
     \end{subfigure}
    
    \caption{Twitter images classified by MemeTector: (a) through (d) are correct predictions, while (e) through (h) are wrong predictions. (a), (b), (g) and (h) are image memes, while (c), (d), (e) and (f) are regular images. The ground truth label is presented in the left of the arrow, and the MemeTector prediction on the right.}
    \label{fig:twitter}
\end{figure*}

Figure~\ref{fig:attention} illustrates attention plots from MemeTector's trainable attention mechanism. Given that this mechanism attends back to four layers, here we show the average attention weights across these four layers. As we observe in image memes the MemeTector attends mostly on the areas where text is present almost ignoring the background content. However, it does not attend at the whole sentences but only at a few seemingly random parts of them which means that analysing the font morphology (being the same throughout each sentence) provides sufficient information for accurate discrimination. Similarly, humans do not need to focus on every part of image memes that contains text to make an informed and accurate decision. In regular images, MemeTector focuses on the main depicted concepts as well as the text if it is present and prominent. Presumably, the reason for not classifying regular images with prominent text presence as image memes is the fonts' morphology 
and their position.

\subsection{Use case on Twitter images}\label{subsec:twitter}

We also evaluate MemeTector on images from Twitter in order to assess its applicability on a practical use case. Specifically, we consider three relevant queries, namely ``meme'', ``dankmemes'', ``memesdaily'' and download 19,502 recent tweets, on 15 April 2022. Out of the collected tweets, only 6,256 contain an image: 2,071 from ``meme'', 1,660 from ``dankmemes'' and 2,525 from ``memesdaily'' query respectively. We download these images, drop the duplicates leaving 3,199 images for analysis and provide them as input to the MemeTector. Another seven of the downloaded images being gray-scale are also discarded because our model has been trained on RGB images only.

The model detected 1,342 memes (42\%) and 1,850 regular images (58\%). To provide quantitative results, we manually labeled the Twitter images, compared with MemeTector's predictions and assessed the model's accuracy. We found TP=877, FP=396, TN=1,454 and FN=465, which amount to a 73\% accuracy. Note that although the used queries are related to image memes there are regular images retrieved. The performance is reduced in the uncontrolled and noisy real-world data as expected but it can still be considered successful especially when considering the quite different characteristics of the images used for training the model. 

In Figure~\ref{fig:twitter}, we illustrate indicative correct and erroneous predictions from the experiment on Twitter images. The correctly classified image meme in Figure~\ref{fig:twitter}a has a similar format to MemeTector's training samples, while the format of Figure~\ref{fig:twitter}b with text over objects or persons is not present in the training set but the model still recognizes it from the fonts morphology and background image semantics. There are many other meme formats not present in the training set that the model recognizes as well. The correctly classified regular images do not confuse the model,  even though they contain text. This is due to the incorporation of text containing regular images at training. The misclassified regular images, contain meme-like overlay text and text at the top, which appears to mislead MemeTector. The first misclassified image meme (Figure~\ref{fig:twitter}g) contains only two numbers as overlay text which is not a common meme form in the training set. However, the second (Figure~\ref{fig:twitter}h) has similar structure to the training samples and MemeTector focuses on the fonts, but their morphology is different and that might be the reason for the detection miss.

\section{Conclusions}\label{sec:conclusions}
In this work, we address the problem of image meme detection. We introduce a novel artificial dataset creation process termed Visual Part Utilization (VPU) that first extracts the visual part of an image meme and then utilizes this new image as an instance of the regular images class. Additionally, we propose a trainable attention mechanism on top of a ViT architecture combining different levels of information granularity that led not only to improved performance but also to interpretability of the model's choices. The findings show that our model surpasses state of the art performance and also demonstrate the usefulness of incorporating VPU in the training of MemeTector. Finally, we validated the proposed methodology on a practical use case involving the retrieval and classification of image memes from Twitter.

\section*{Declarations}
\textbf{Competing interests} The authors have no competing interests to declare that are relevant to the content of this article.

\noindent
\textbf{Funding} This work is partially funded by the Horizon 2020 European project MediaVerse under grant agreement no. 957252.

\noindent
\textbf{Data availability.} All datasets used in this work are publicly available and have been properly referenced in the text.

\bibliography{IJMIR/sn-bibliography}
\bibliographystyle{unsrt}
\end{document}